# PREDICTIVE REPAIR MANAGEMENT USING A MULTI-HEAD ATTENTION TRANSFORMER AND ONLINE LEARNING

**Xinyao Zhang**
University of Florida
Gainesville, FL, USA
xinyaozhang@ufl.edu

**Willie Cade**
Graceful Solutions
Chicago, IL, USA
willie@graceful.solutions

**Karl R. Haapala**
Oregon State University
Corvallis, OR, USA
karl.haapala@oregonstate.edu

**Arun Natarajan**
Oregon State University
Corvallis, OR, USA
arun.natarajan@oregonstate.edu

**Sara Behdad***
University of Florida
Gainesville, FL, USA
sarabehdad@ufl.edu

## ABSTRACT

*Accurate prediction of repair duration is an important challenge in product maintenance due to its implications for resource allocation, customer satisfaction, and operational performance. This study aims to develop a deep learning framework to help fleet repair shops accurately categorize repair time given product historical data. The study uses an automobile repair and maintenance dataset and creates an end-to-end predictive framework by employing a multi-head attention network designed for tabular data. The developed framework combines categorical information, transformed through embeddings and attention mechanisms, with numerical historical data to facilitate integration and learning from diverse data features. A weighted loss function is introduced to overcome class imbalance issues in large datasets. Moreover, an online learning strategy is used for continuous incremental model updates to maintain predictive accuracy in evolving operational environments. Our empirical findings demonstrate that the multi-head attention mechanism extracts meaningful interactions between vehicle identifiers and repair types compared to a feed-forward neural network and a random forest model. Also, combining historical maintenance data with an online learning strategy facilitates real-time adjustments to changing patterns and increases the model's predictive performance on new data. The model is tested on real-world repair data spanning 2013 to 2020 and achieves an accuracy of 78%, with attention weight analyses illustrating feature interactions.*

## 1. INTRODUCTION

Lifecycle data provides a complete record of a product's history, including its design, manufacturing details, operational performance, failure events, and repair records. This information is valuable for making end-of-use decisions, optimizing maintenance schedules, and guiding repair strategies [1]. Also, lifecycle data helps assess repair complexity and predict the time required to service different products. Since lifecycle data is generated throughout the entire product lifespan, the data might have diverse formats, such as numerical values, textual records, and sensor-related data [2]. This heterogeneity creates a wide range of analytical and functional applications. Variations in lifecycle data, including information about the product design, component accessibility, and historical failures, lead to different repair decision patterns and influence repair duration [3].

Lifecycle data have been used in maintenance and reliability studies to support failure diagnosis, reliability assessment, preventive maintenance planning, and remaining useful life estimation. Prior work has mainly employed lifecycle information to model degradation behavior and system health over time. Although such studies have advanced understanding of system behavior, much of the existing work has focused on failure prediction or reliability assessment rather than on estimating repair duration after a failure occurs. In practice, accurate repair duration estimates are important for maintenance planning, workforce scheduling, and service coordination, particularly in automotive repair environments where vehicle configurations, repair procedures, and operating conditions change over time. These considerations motivate the development of data-driven methods that can utilize heterogeneous lifecycle data to support repair duration prediction.

The objective of this work is to formulate and address the problem of repair duration prediction using heterogeneous lifecycle data associated with failure and repair events. The problem is characterized by high-dimensional tabular inputs, mixed data types, and variability between repair instances arising from differences in vehicle attributes and historical maintenance records. The objective is to develop a modeling approach that represents interactions among these features and maintains predictive performance as new repair data becomes available.

To achieve this objective, this research makes the following contributions:

- The TabTransformer architecture is applied to vehicle repair duration prediction, a domain where attention-based methods have not been explored. The model generates dense representations of vehicle attributes through embeddings and uses self-attention layers to learn feature interactions.
- Online learning is incorporated with the pre-trained Transformer for incremental parameter updates via mini-batch gradient descent. This combination facilitates real-time adjustment to new repair patterns without catastrophic forgetting.
- Domain-specific feature engineering is designed including VIN embeddings that model vehicle-specific repair propensities and historical features that encode maintenance patterns over time.

The remainder of this paper is organized as follows. Section 2 introduces the gaps and motivation of study. Section 3 describes the dataset and feature engineering process. Section 4 presents the methodology, including the multi-head attention Transformer architecture and online learning strategy. Section 5 discusses the experimental results and analysis, and finally Section 6 concludes the paper and outlines future work.

## 2. BACKGROUND

The concept of repairability has gained significant attention in recent years, especially with the growing influence of the Right to Repair movement [4]. Although reliability has long received attention in the literature, it differs from repairability. Reliability concerns estimating product lifespan or predicting failure time [5], whereas repairability refers to the ease of maintenance and restoration after failure. Repair data has been used to estimate the time between failures and schedule preventive maintenance, but its applications extend beyond failure prediction [6]. It also serves as an essential measure of product longevity and design performance.

To quantify these aspects, researchers have developed repairability indices [7], which assess how factors such as modular design [8], ease of disassembly [9], and spare part availability influence repair costs and turnaround times [10]. Learning models are used to analyze equipment wear and optimize maintenance schedules based on real-time sensor data and historical records [11]. In aerospace, predictive analytics applied to aircraft component lifecycles has improved flight safety and maintenance cost [12]. Other industries, such as healthcare and consumer electronics, have used lifecycle data to improve maintenance planning and reduce downtime [13, 14].

Predictive maintenance has received attention in the literature, however, repair duration prediction is still an underexplored area. Prior studies have developed predictive maintenance techniques that integrate lifecycle data with machine learning to forecast failure events [15], estimate remaining useful life [16], and optimize maintenance intervals [17]. However, much of the previous research has focused on predicting failures before they happen; less attention has been given to estimating repair durations and how long it takes to fix a product once a failure occurs. This study addresses that gap using machine learning to improve repair time predictions.

To optimize automative maintenance management, the research motivation is to achieve accurate prediction of repair durations. Precise forecasts directly affect resource allocation, reduce operational costs, and improve customer satisfaction by minimizing service delays [18]. However, achieving reliable predictions is challenging due to dynamic factors such as heterogeneous data types (e.g., categorical attributes and numerical maintenance histories) and temporal shifts in repair patterns [19, 20]. Factors such as class imbalance when categorizing repair durations into discrete intervals can also affect forecasting [21]. Thus, solutions should process multi-feature data and handle evolving workflows.

Existing approaches for repair duration prediction span statistical, ensemble, and deep learning methods. Classical techniques, including survival analysis and linear regression, model repair times as time-to-event or continuous outcomes, but often underperform with high-dimensional categorical data [22]. Ensemble methods such as XGBoost and LightGBM address these limitations by handling nonlinear feature interactions, but their static training paradigms limit interpretability and applicability in streaming data scenarios [23-25]. Recent advances in deep learning, particularly Transformer-based architectures, offer promising improvements. Models such as TabTransformer leverage self-attention mechanisms to contextualize categorical embeddings [26]. Another model named TabNet designs sequential attention that achieves state-of-the-art accuracy in tabular data tasks while providing interpretable feature attributions [27]. Nevertheless, existing frameworks do not incorporate mechanisms for continuous model updates, a critical gap in maintenance where data distributions shift over product fleets and repair technologies.

Besides seeking to predict repair durations accurately from existing data, it is necessary to preserve model performance as new data arrive. Online learning refers to a machine learning approach where models update as new data becomes available, rather than training once on a static dataset [28]. For repair shops, online learning strategy is valuable as it helps prediction models adjust to patterns in repair complexity, parts availability, and technician expertise [29]. In automotive repair, online learning helps maintenance facilities improve operations by providing accurate time estimates based on real-world repair outcomes. This research follows an input-update model, in which new repair records are continuously fed into the system to modify predictions [30]. This strategy is designed for repair shops with varying workloads and changing vehicles to maintain model relevancy without requiring complete retraining.

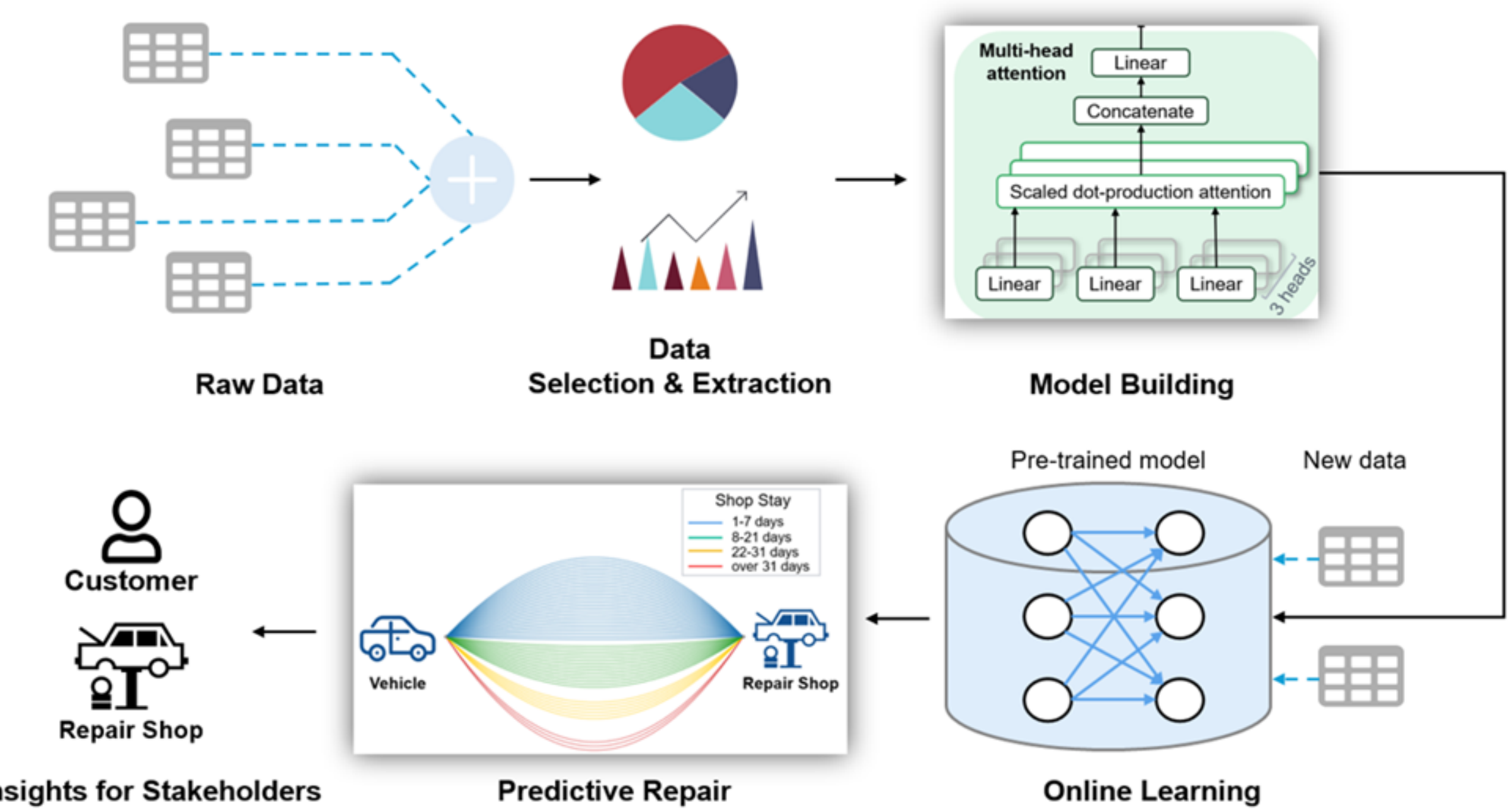

FIGURE 1: PREDICTIVE REPAIR MANAGEMENT DATA FLOW

This research develops an end-to-end attention-based model with integrated online learning for repair duration classification. Figure 1 illustrates the resulting framework for predicting the repair duration. The framework comprises raw data processing, model building, prediction, and online learning. Initially, raw data are integrated and go through preprocessing to generate structured inputs. These inputs are fed into a multi-head attention model that utilizes scaled dot-product attention to extract complex interactions between features. Then, an online learning mechanism incrementally updates the pre-trained model with incoming data to maintain high accuracy and flexibility in real-world scenarios. Finally, the predictive outcomes, classified as duration-based repair categories, provide guidelines for stakeholders such as customers and repair shops.

## 3. DATASET AND FEATURE ENGINEERING

The data consists of 9,103 repair records for vehicles produced by the same manufacturer. Table 1 provides an overview of the dataset, which includes six main attributes: vehicle identification (VIN), production year, model, repair description, the start and end dates of each repair event, and cost.

Since there are limited features present in the dataset, we use feature engineering to extract and compute additional features that complement the information. Table 2 explains the extracted features. The Age feature is calculated as the difference between the vehicle's production and repair years. Number of Repairs tracks how many prior repairs the vehicle underwent before the current event. Time Between Repairs measures the time elapsed since the last recorded repair.

**TABLE 1:** DATASET EXPLANATION

| Feature Name | Description | Characteristic |
|---|---|---|
| VIN | Vehicle identification number | 531 unique values |
| Year | Vehicle production year | Formatted as year |
| Model | Vehicle model | 10 unique values |
| Repair Description | Description of the repair event | 12 unique values |
| Start Date | Start date (time) for a repair event | Formatted as month/day/year hour:minute:second |
| End Date | End date (time) for a repair event | Formatted as month/day/year hour:minute:second |
| Cost | Total repair cost | Numeric value |

Repair Duration is a categorical variable that assigns each repair event to a time-interval group. The Repair Duration feature represents the length of the repair and the time a vehicle remains at the repair shop, from arrival to repair completion. The Start Date and End Date fields contain timestamp information that includes both date and time, allowing repair duration to be calculated with hour-level precision. For single-day repairs, we subdivide the duration into hour-level categories to address class imbalance. Without this subdivision, the majority of records would fall into a single category, which reduces the model's practical utility. For multi-day repairs, we use day-level categories. This 6-class formulation provides actionable categories for repair shops to estimate turnaround times and allocate resources.

**TABLE 2:** EXTRACTED FEATURES

| Feature Name | Description | Characteristic |
|---|---|---|
| Age | Year of repair minus year of production | Formatted as whole years, starting at 1 |
| Number of Repairs | Number of prior repairs the vehicle had before the current one | Formatted as integers |
| Time Between Repairs | Days since the vehicle's last repair | Formatted as whole days |
| Repair Duration | Categorized repair time | 6 unique groups |

The selected input features include VIN, Model, Repair Description, Cost, Age, Number of Repairs, and Time Between Repairs. The output, termed as Repair Duration, shows the time a vehicle is at a repair facility to complete repair or maintenance tasks. The proposed model extracts interactions among input features to predict vehicle Repair Duration accurately. Our objective is to predict vehicle Repair Duration using categorical vehicle characteristics and historical maintenance data. For modeling purposes, Repair Duration is categorized into six separate groups, including 1 hour, 1-2 hours, 2-4 hours, over 4 hours in 1 day, 2-7 days, and over 7 days.

The main characteristics of the dataset are as follows. VIN serves as a unique identifier for each vehicle. There are 531 unique VINs, and many vehicles have multiple repairs recorded, resulting in an average of 26 repairs per vehicle. Types of Repair Descriptions indicate the nature of repair activities by specifying which components are repaired or serviced. These descriptions are categorized into hydraulics, preventative maintenance, and others. A higher count could show an older or trouble-prone vehicle component, potentially affecting repair difficulty. The number of repairs reveals the chronological nature of the data. A higher count could also indicate an older or trouble-prone vehicle. A short interval since the last repair might mean unresolved issues or recent maintenance means less work now. For categorical features, instead of one-hot encoding, we transform categorical features into learned embeddings.

Each VIN gets an embedding in 64 dimensions to capture its unique propensity for repair durations. We embed the vehicle Model in 32 dimensions. This represents differences such as trucks versus sedans or model-specific reliability. The Repair Description is embedded in 32 dimensions. Historical features are dynamic per record and give the model a temporal setting. We normalize continuous historical data using Z-scores [31]. This way, no feature dominates due to scale. For example, costs range in hundreds of dollars and do not overshadow hours, which are single digits simply due to magnitude.

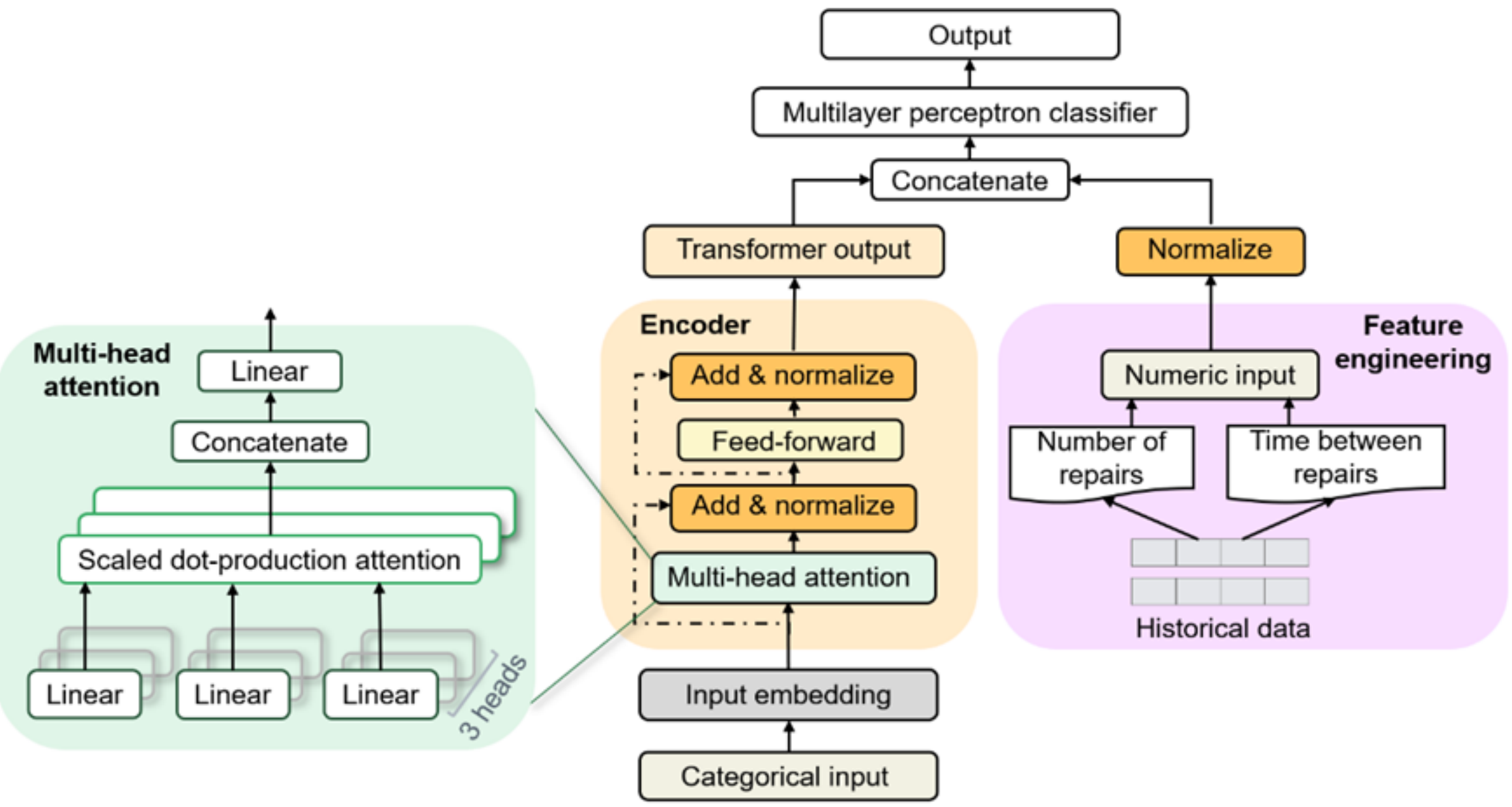

**FIGURE 2:** MULTI-HEAD ATTENTION TRANSFORMER ARCHITECTURE

We combine static categorical information with dynamic historical features to give the model a detailed view of each repair record.

## 4. METHODOLOGY

Our model draws inspiration from the Transformer's encoder module [32], illustrated in Figure 2. The multi-head self-attention mechanism is at its core, which helps the model learn interactions between input features.

### 4.1 Multi-Head Attention Transformer

Self-attention is designed initially in natural language processing to let a sequence model weigh the relevance of different words in a sentence to each other. Here, we apply it to the sequence of feature embeddings. The idea is to let the model learn, for each feature, how much attention to pay to other features when predicting the output. For example, embedding the Repair Description might pay attention to the VIN embedding if the vehicle's identity influences that specific repair event.

Given queries $Q$, keys $K$, and values $V$, which in self-attention are all derived from the input embeddings matrix $X$ [32], the attention output is (Eq. 1):

$$Attention(Q, K, V) = \text{softmax}(\frac{QK^T}{\sqrt{d_k}})V \tag{1}$$

where $d_k$ is the dimensionality of the key vectors. In our tabular application, $X$ represents the concatenated embeddings of categorical features (i.e., VIN, Model, and Repair Description), where each feature embedding serves as a token in the attention computation.

Multi-head attention extends this by having multiple sets of $W$ matrices, i.e., multiple attention heads, so the model can extract different types of relationships in parallel. Each head $i$ computes its attention output, and the heads are concatenated and linearly transformed to form the final output. If we use $h$ heads, each with key dimensionality $d_k$ and value dimensionality $d_v$, then (Eqs. 1-2):

$$MultiHead(X) = [head_1, \dots, head_h]W^o \tag{2}$$

$$head_i = Attention(XW_i^Q, XW_i^K, XW_i^V) \tag{3}$$

where $W^o$ is an output projection and $X$ is a matrix containing the embeddings of categorical features for a single record.

The structure of the model is displayed in Figure 2. We develop a deep learning architecture, TabTransformer [26]. TabTransformer leverages Transformer-based self-attention to convert categorical feature embeddings into contextual embeddings for both supervised and semi-supervised tasks. Applying this approach improves data interpretability. Our contribution is not architectural innovation but rather the novel application of this framework to repair duration prediction with three key adaptations: domain-specific embedding design for vehicle and repair features, incorporation of online learning for temporal deployment, and weighted loss calibration for imbalanced repair categories.

Categorical inputs, such as Model and VIN, are turned into embeddings. Since the features are distinct, a fixed embedding per feature can be added to its vector to indicate the feature identity. A Transformer encoder block consists of multi-head attention and a feed-forward network. This produces contextualized representations for each categorical feature. We apply a feed-forward network to each embedding, consisting of two dense layers, as in Transformers. Residual connections and layer normalization are connected with the multi-head attention. The outputs of the Transformer for each feature are concatenated into one vector. We also append the normalized numeric feature vector. This combined vector goes through a multi-layer perceptron containing dense layers with Gaussian Error Linear Unit (GELU) activation and, finally, a SoftMax layer that outputs probabilities for each duration group.

The model uses embedding dimensions of 64 for VIN, 32 for Model, and 32 for Repair Description. The Transformer encoder consists of 4 stacked blocks, each with 8 attention heads and a key dimension of 32. The classification head contains three fully connected layers with dimensions 128, 64, and 32.

### 4.2 Online Learning

Online learning constitutes a class of machine learning approaches wherein a model addresses predictive or decision-making tasks by sequentially learning from individual data instances as they become available [28]. The objective is to improve prediction or decision accuracy based on knowledge acquired from prior outcomes and supplementary information. This approach differs from traditional batch or offline machine learning techniques, which train models using entire datasets simultaneously. Online learning is also different from fine-tuning. Fine-tuning refers to updating a pre-trained model on a specific dataset to adapt it to a particular task. This includes a one-time adjustment of model weights to optimize performance, but online learning is a continuous model adaptation paradigm. Online learning has gained more attention due to its applicability for handling continuous data in various real-world applications.

We used a standard supervised online learning strategy, as illustrated in Figure 3, with hyperparameter customizations described in Section 5.5 to preserve learned embeddings during incremental updates. Specifically, the pre-trained model initially predicts over the first four months of 2021 and then incorporates ground truth data through fine-tuning to improve its predictive accuracy. This iterative updating maintains high accuracy when forecasting for the next four months of 2021.

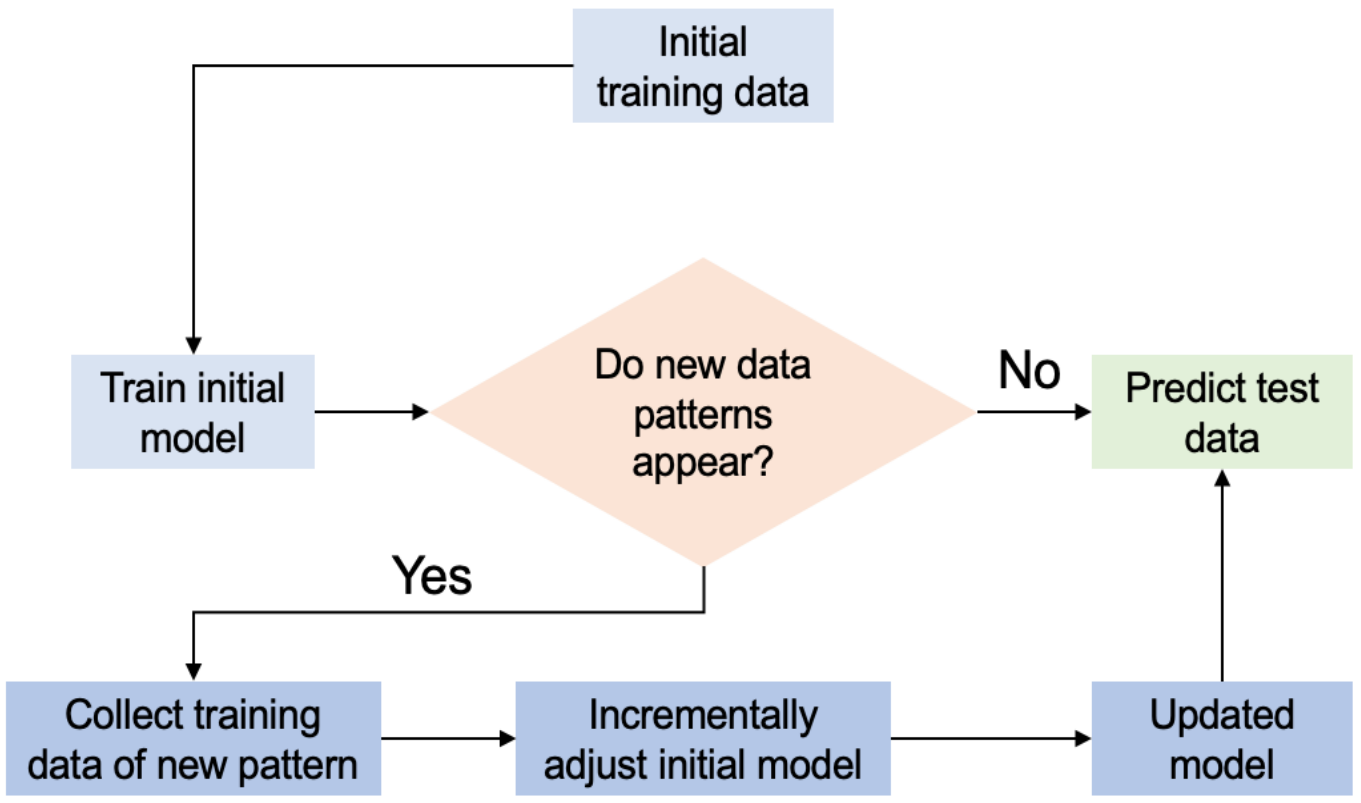

**FIGURE 3:** THE FLOW CHART OF ONLINE LEARNING

When using online learning, we should be cautious to prevent the model from forgetting past patterns and overfitting recent data. One strategy is to train the model for a few epochs on new data with a lower learning rate. We use a small batch size to preserve past knowledge, reduce overfitting, and limit the number of training epochs.

## 5. RESULTS AND DISCUSSION

This section presents the results in detail and interprets each finding. First, we discuss data preprocessing and model training. Then, we present the Transformer's prediction results.

### 5.1 Data Preprocessing and Model Training

The pre-training dataset consists of 7,861 records spanning from 2013 to 2020, which are partitioned into 85% training and 15% validation sets. The target distribution for Repair Duration is around 41% for 1 hour, 13% for 1-2 hours, 13% for 2-4 hours, 8% for over 4 hours in 1 day, 14% for 2-7 days, and 11% for more than 7 days. We use stratified sampling to keep the target distribution consistent between the training and validation sets. Data from 2021 is set aside to test incremental learning.

The model is trained using categorical cross-entropy loss with inverse frequency class weighting. We use the Adam optimizer with a learning rate of 0.001 and batch size of 32. Training runs for up to 100 epochs with early stopping based on validation loss to prevent overfitting. During training, precision and recall for minority classes are monitored to verify that the model detects them correctly.

### 5.2 Results of Multi-Head Attention Transformer

We evaluate the model by classifying repairs based on their duration. The model achieves an overall accuracy of 77.88% on the validation set. Table 3 lists the detailed predicted accuracy per class, and Figure 4 shows the confusion matrix.

The class-specific results in Table 3 demonstrate the Transformer's ability to capture complex feature interactions. The model achieves 100% accuracy for the "Over 4 hours in 1 day" class and 94.7% for "Over 7 days," which are minority classes comprising only 8% and 11% of the dataset, respectively. This strong performance on underrepresented categories indicates that the attention mechanism learns meaningful patterns from feature interactions. Specifically, the model captures relationships among vehicle identifiers, repair types, and historical features that are critical for predicting extended repair durations. Recent benchmark studies on tabular data classification support these findings. Gorishniy et al. [33] and McElfresh et al. [34] demonstrate that transformer architectures can match or exceed tree-based methods when structure-aware representations of categorical features benefit prediction.

**TABLE 3:** VALIDATION PREDICTION ACCURACY

| Repair Duration | Accuracy |
|---|---|
| 1 hour | 0.771 |
| 1-2 hours | 0.677 |
| 2-4 hours | 0.814 |
| Over 4 hours in 1 day | 1.000 |
| 2-7 days | 0.636 |
| Over 7 days | 0.947 |
| **Overall** | **0.779** |

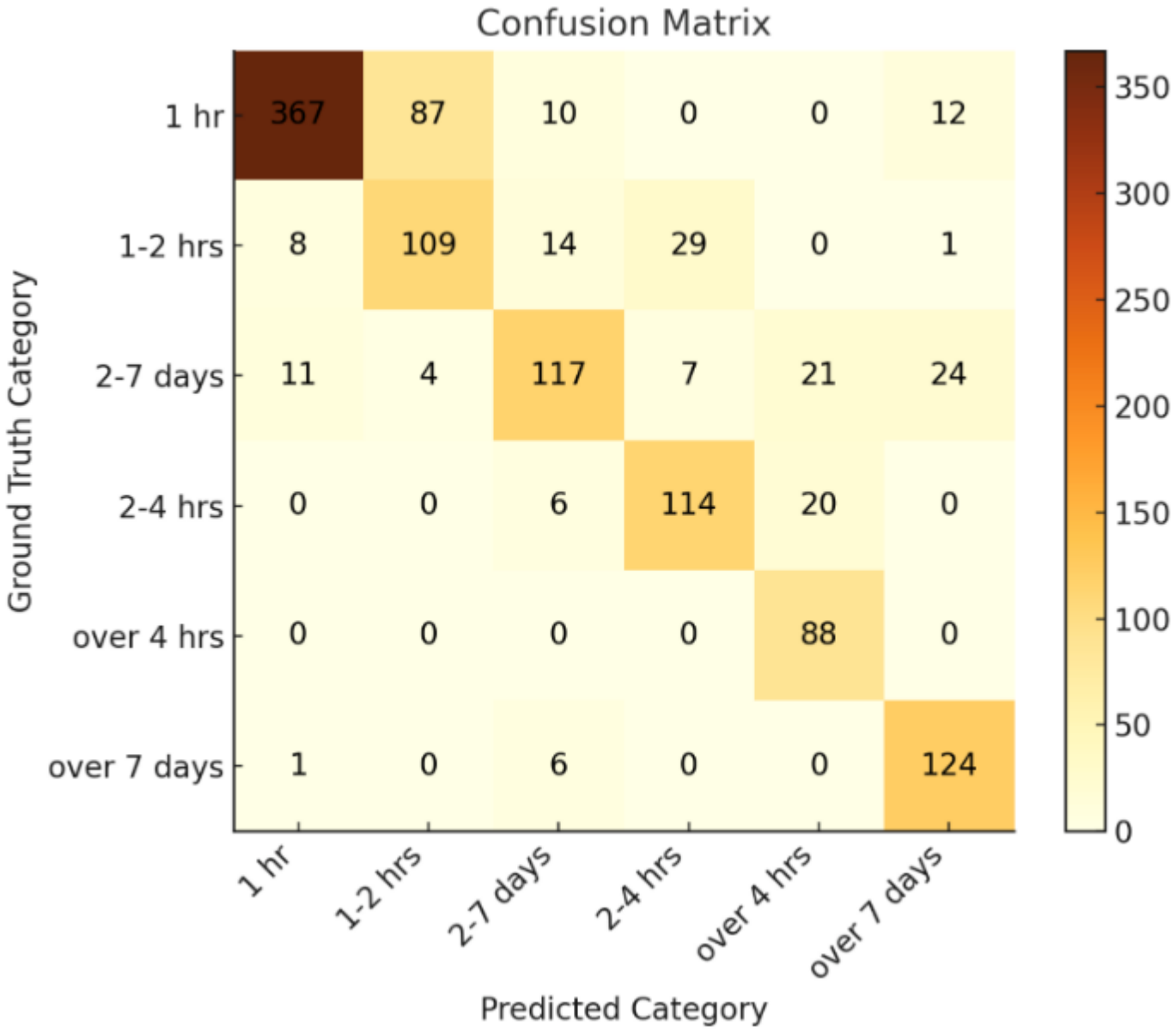

**FIGURE 4:** TRANSFORMER VALIDATION PREDICTION CONFUSION MATRIX

### 5.3 Comparison Results of Feed-Forward Neural Network and Random Forest

To benchmark the Transformer, we compare it against two baselines: a feed-forward neural network that mirrors the Transformer's dense blocks, and a random forest as a classical ensemble reference. The feed-forward model maintains architectural components with the Transformer's feed-forward sublayers, while the random forest provides a non-deep-learning point of comparison. Quantitative evaluation is displayed in Table 4.

**TABLE 4:** MODEL ACCURACY COMPARISON

| | Transformer | Feed-Forward Neural Network | Random Forest |
|---|---|---|---|
| **Accuracy** | **77.88%** | 76.69% | 71.36% |

The feed-forward model utilizes identical input features but differs in architectural complexity. It employs simple embedding layers followed by dense connections rather than the Transformer's self-attention mechanisms and positional encodings. The model achieved an accuracy of 76.69% on the validation set, while the Transformer reached a higher accuracy of 77.88%. The confusion matrix for the feed-forward model is displayed in Figure 5.

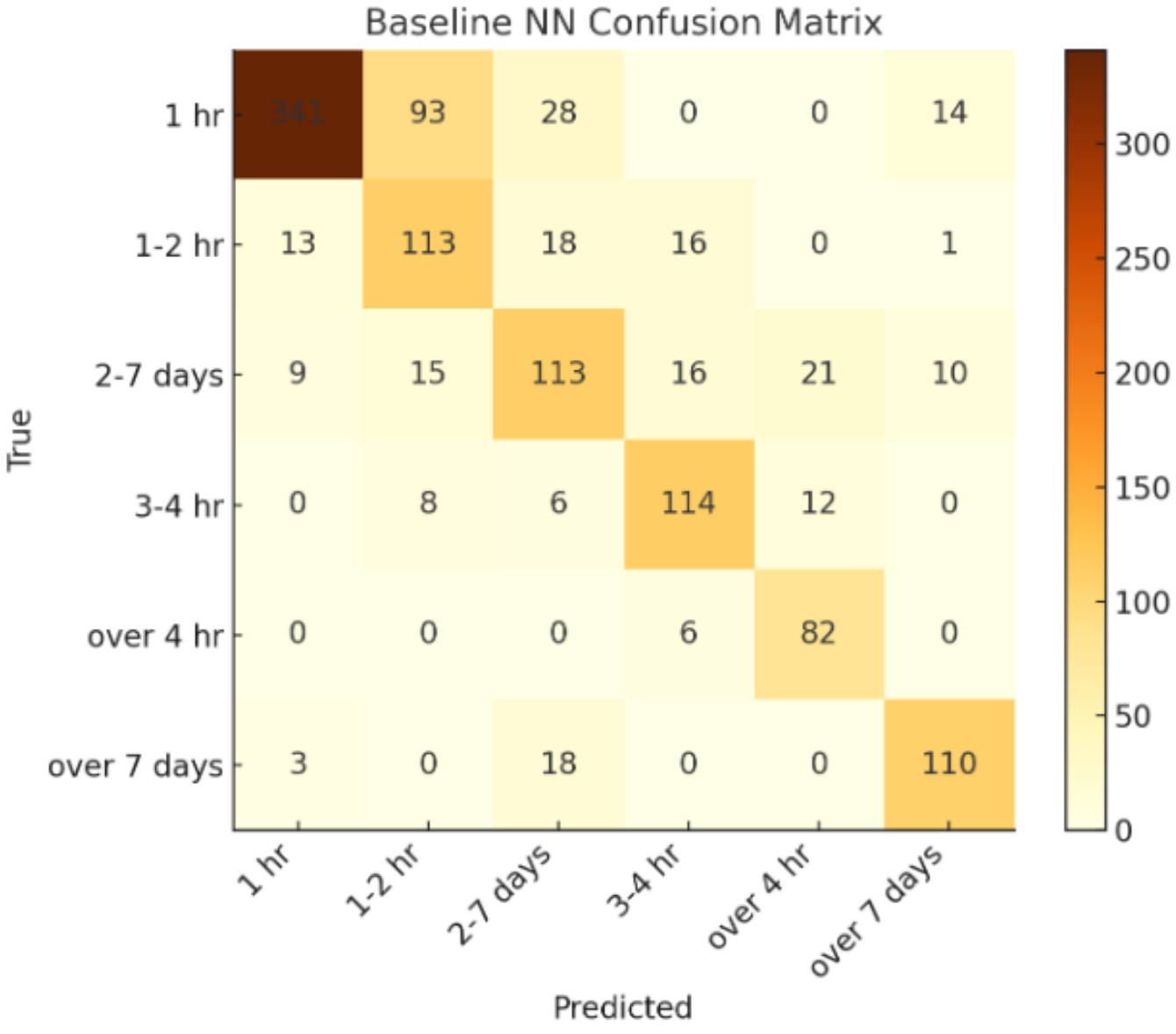

**FIGURE 5:** FEED-FORWARD NEURAL NETWORK VALIDATION PREDICTION CONFUSION MATRIX

The Transformer's performance can be attributed to three main factors: its attention mechanism's ability to model complex interdependence between repair attributes, its capacity to learn contextual representations of categorical variables through embedding transformations, and its deeper architecture to capture higher-order patterns in the temporal aspects of vehicle repairs. These architectural advantages are valuable for modeling the sequential and relational nature of repair duration prediction tasks.

To further assess model performance, we include a random forest baseline. Random forests handle categorical features without the need for embedding transformations, and their nonlinear decision boundaries can capture patterns in tabular data. In our experiments, the Transformer outperforms both the feed-forward neural network and the random forest, achieving 77.88% accuracy compared to 76.69% and 71.36%, respectively. This improvement suggests that the attention mechanism captures meaningful interactions among vehicle identifiers, repair types, and historical features that tree-based models cannot fully represent. Furthermore, the Transformer's architecture supports online learning for continuous implementation, which is valuable for maintaining prediction accuracy as repair patterns evolve over time.

### 5.4 Results of Attention Mechanism

The multi-head attention mechanism provides some interpretability. We examine the attention weight matrices for validation examples presented in Table 5. For many samples of long repairs, the VIN and Historical Features receive strong attention, which suggests that the model considers the combination of vehicle identity and past repair events when identifying long repairs. Meanwhile, the VIN appears as the dominant feature and shows the importance of vehicle-specific characteristics in determining repair duration.

**TABLE 5:** MULTI-HEAD ATTENTION IMPORTANCE

| | Model | Repair Description | VIN | Historical Features |
|---|---|---|---|---|
| **Attention Importance** | 3.8% | 3.9% | 85.6% | 6.7% |

Both Model and Repair Description features exhibit minimal influence. Model embedding often has lower attention weights, except in cases where the vehicle model is unusual in repair duration. The numeric features, such as Age, Number of Repairs, and Time between repairs, consist of Historical Features. They reveal previous repair patterns and temporal information.

Overall, the attention weights indicate that the model relies primarily on vehicle-specific characteristics captured through the VIN embedding when predicting repair duration. The attention mechanism explicitly models pairwise relationships. This interaction-based reasoning explains the Transformer's superior performance on minority classes in Table 3, where understanding relationships between specific vehicles and their repair patterns is critical.

### 5.5 Results of Online Learning

To maintain performance over time, we periodically retrain or fine-tune the model as new data comes in, such as quarterly or semi-annually. This helps the model respond to changes in new vehicle models, aging, or changing maintenance practices.

Applying the Transformer model pre-trained in 2013-2020, we first fine-tune the model using recent observations in January-April 2021 and later predict repair durations for subsequent months in May-August 2021. The implementation uses feature engineering with temporal and historical indicators and strategic hyperparameter optimization, including learning rate and early stopping protocols. For online learning, we reduce the learning rate to 0.0001 to preserve learned representations while adapting to new patterns. We use a batch size of 16 and train for 5 epochs per update cycle. These conservative settings help prevent catastrophic forgetting of the vehicle-specific embeddings learned during pre-training.

Results achieve performance improvement of 74.3% in prediction accuracy, from 19.55% to 34.08%, following online learning implementation. The results show the value of online learning techniques in mitigating model drift within temporal prediction systems. Class-specific metrics demonstrate gains in extreme duration categories, with accuracy improvements of 50.90% for 1-hour repair durations and 32.61% for repair durations of over 7 days. This performance improvement can be attributed to factors such as adaptation to temporal data drift patterns, restoration of classification capability in previously undetected classes, and improved calibration over the prediction spectrum. However, the absolute accuracy of 34.08% remains low, which indicates that conservative update strategies are insufficient for distribution shifts.

## 6. CONCLUSION

This study developed an end-to-end predictive framework for classifying repair durations using a multi-head attention Transformer with online learning. The model processes categorical and numerical historical data to identify the meaningful relationships between vehicle identifiers, repair types, and maintenance history. A weighted loss function has been applied to address class imbalance in the dataset. Also, an incremental learning strategy has been used to help the model adjust to changes in repair trends over time.

The model is trained on a dataset of vehicle repair records from 2013 to 2020, covering various repair categories and durations. It achieves 78% accuracy on the validation dataset. Analysis of attention weights demonstrates that vehicle identity is the most influential factor, which indicates that vehicle-specific characteristics and historical patterns play a dominant role in determining repair duration. The model performs well; however, it has some limitations. The Transformer model might be overparameterized for this dataset size. The performance of the Transformer model can be improved by modifying the model structure and designing more extensive hyperparameter tuning. Also, the pre-trained model has limited performance on the unseen data.

The work can be extended in several ways. Future work will investigate self-supervised learning to improve feature extraction from limited data and uncertainty quantification to assess confidence in predictions. Hybrid models that combine deep learning with traditional machine learning methods may also increase accuracy. Future work will expand our dataset to include repair records from more manufacturers, electric and hybrid vehicles, and diverse regions to improve generalizability. Moreover, the emergence of smart vehicles facilitates the implementation of in-vehicle diagnostics and Internet of Things (IoT) sensors to build large, real-time datasets. Lightweight models at the edge can provide instant duration estimates, while cloud-based architectures such as Transformers, learn from IoT streams. Another direction of future work is to improve online learning performance through replay-based methods that mix historical examples with new data, layer-specific learning rates that protect learned embeddings, and more frequent update cycles to capture temporal drift earlier. Finally, incorporating real-time feedback from repair shops could optimize predictions and make the system more practical.

## ACKNOWLEDGEMENTS

This material is based upon work supported by the U.S. National Science Foundation under Award Nos. 2324950, 2324949, and 2412471. Any opinions, findings, conclusions, or recommendations expressed in this material are those of the authors and do not necessarily reflect the views of the National Science Foundation.